\documentclass{article}
\pdfoutput=1
\usepackage{spconf,amsmath,graphicx,color}
\usepackage{booktabs}

\title{SCALABLE MULTILINGUAL FRONTEND FOR TTS}
\name{Alistair Conkie, Andrew Finch}
\address{Apple}

\begin{document}
\maketitle
\begin{abstract}
This paper describes progress towards making a Neural
Text-to-Speech (TTS) Frontend that works for many languages and can be
easily extended to new languages.
We take a Machine Translation (MT) inspired approach to constructing the
frontend, and model both text normalization and pronunciation on a
sentence level by building and using sequence-to-sequence (S2S)
models. We experimented with training normalization and pronunciation as
separate S2S models and with training a single S2S model
combining both functions.

For our language-independent approach to pronunciation we do not use a
lexicon. Instead all pronunciations, including context-based
pronunciations, are captured in the S2S model.
We also present a language-independent chunking and splicing
technique that allows us to process arbitrary-length sentences.
Models for 18 languages were trained and evaluated.
Many of the accuracy measurements are above 99\%.  We also evaluated the
models in the context of end-to-end synthesis against our current
production system.

\end{abstract}

\section{INTRODUCTION AND RELATED WORK}
\label{sec:introduction}

Text-to-Speech synthesis has made tremendous progress over the last
twenty or so years, above all in terms of naturalness of the output
voice. For an overview see \cite{taylor_2009}. Most recently synthesis
quality has improved due to innovative Machine Learning (ML)
techniques such as WaveNet \cite{DBLP:journals/corr/OordDZSVGKSK16}.
It is relatively straightforward to apply these approaches to
different languages.

There has also been steady progress in terms of the frontend (FE) --
normally considered to be the part of a TTS system that converts input
text to a phonetic representation.  There are several reasons why this
is a harder problem than backend waveform generation.  In part it is
because designing a frontend is an intrinsically more knowledge-based
task.  Some successful examples of multilingual synthesis include
\cite{hertz:multilingual}, \cite{sproat:multilingual},
\cite{sproat:1996}.

For a number of years Weighted Finite State Transducer solutions
(WFSTs) were very popular.  This approach \cite{sproat:1996},
\cite{roark+al+2012} is essentially a sophisticated rule-based
approach. Using WFSTs requires a knowledge of linguistics and
also the ability to write formal grammars that then get compiled into
\mbox{WFSTs }.

More recently, work has focused on more general data-driven ML
approaches. Several recent systems are capable of learning directly
from character input \cite{DBLP:journals/corr/OordDZSVGKSK16},
\cite{DBLP:journals/corr/WangSSWWJYXCBLA17},
\cite{DBLP:journals/corr/abs-1712-05884},
\cite{DBLP:journals/corr/ArikCCDGKLMRSS17},
\cite{Sotelo2017Char2WavES}.  The challenge for End-to-End (E2E)
approaches is to have enough training data to train a high quality
system.

In terms of text normalization see e.g. \cite{DBLP:journals/corr/SproatJ16}, where the authors
propose a system using a large parallel corpus to
train models for various recurrent neural network (RNN)
architectures. They found it necessary to add a FST-based
post-filter to achieve the required accuracy. Other hybrid methods
have also been proposed \cite{pusateri}. In \cite{yolchuyeva}, the
use of convolutional neural networks (CNNs) for text
normalization was examined. In \cite{amazon} the authors examine
aspects of text normalization in the context of MT and using byte pair
encoding (BPE) for subword units.

This paper describes our research, where the main idea is to treat the
whole frontend as one or more S2S tasks in a very general way. We
elaborate on this in the following sections.

\section{SYSTEM}
\label{sec:system}

The aim of our work is to model both normalization and pronunciation
(sometimes called grapheme-to-phoneme (G2P)) to provide all the
information necessary for input to the TTS backend -- e.g. WaveNet.
The input is raw text in an unnormalized form, and the output is a
sequence of phonemes along with some additional information.

We regard the frontend as exactly equivalent to a translation task
and employ the tools of MT directly. The task of building a frontend
can be considered as either one or two translation tasks, depending on
how the problem is structured.

If configured as two tasks we find it convenient to divide the problem
into normalization and pronunciation, since it fits in well with how
the problem is conventionally structured.

To a first approximation local context is most important for
clarifying the normalization of a character sequence, or the
pronunciation of a word.  However, in our case we find it useful to
consider whole sentences (1) for the practical reason that the MT
infrastructure is focused on sentences and (2) there are some long
term dependencies that can guide normalization and pronunciation, for
example related to given vs. new distinctions \cite{hirschberg}.

Pronunciations are also generated from the translation sentence
context, rather than from a lexicon. We use parallel data in the form
of pairs of sentences, with input in the form of normalized words, and
the output in terms of phonemes.  We then train a S2S model based on
that data, and use the model to generate pronunciations for our input
sentences (or words).  For our experiment we use only parallel data,
with no extra helper knowledge, and the only data supplied at training
time is input and output examples (training set and development set).
In the one-model case, the modeling must take account of both sets of
challenges, normalization and pronunciation.

\subsection{Transformer Model Architecture}
\label{subsec:fairseq}

\begin{figure}[htb]
		\begin{minipage}[b]{0.95\linewidth}
			\centering
			\centerline{\includegraphics[width=9cm]{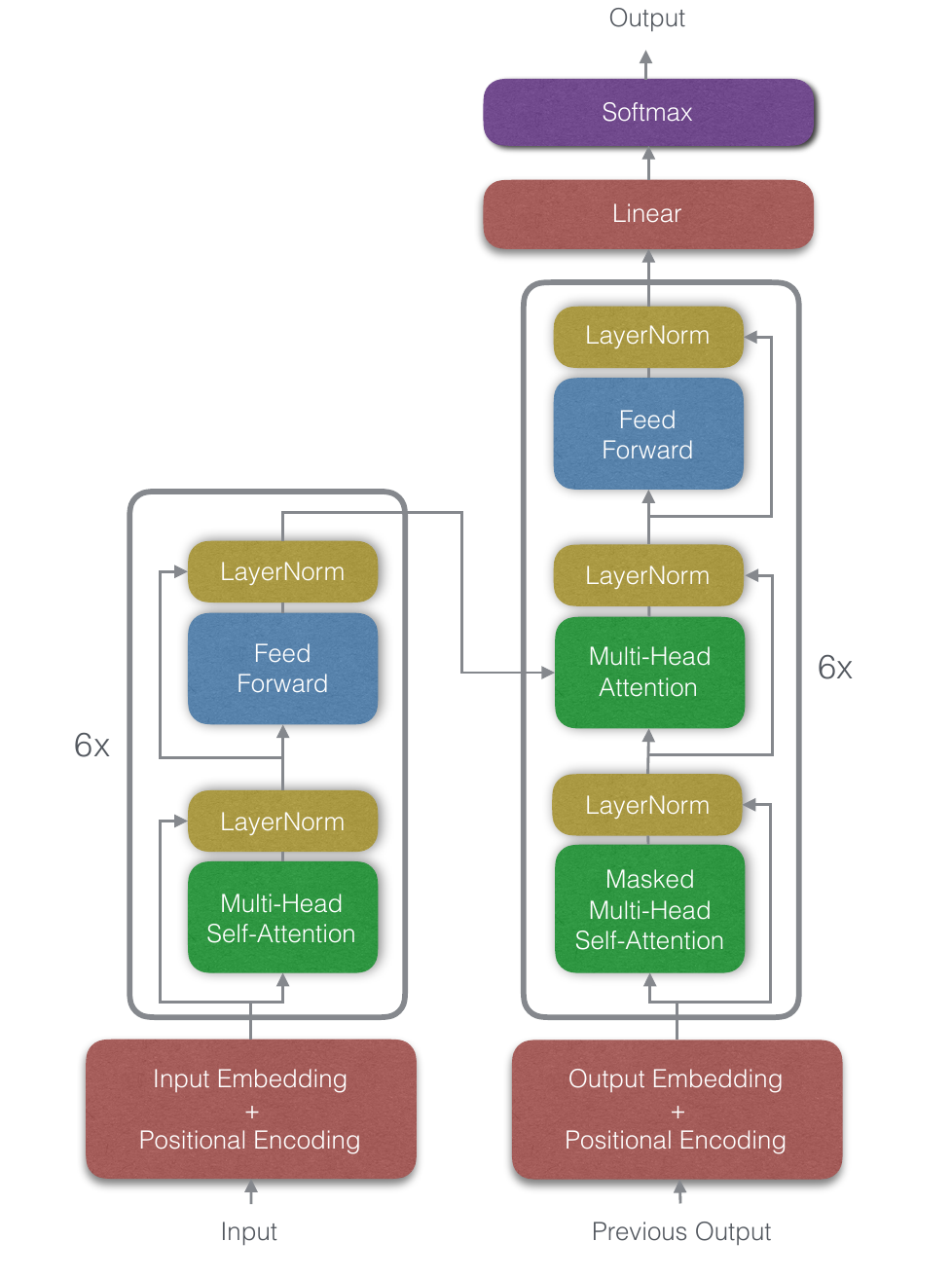}}
		        \vspace{-0.1cm}
		\end{minipage}
		\caption{Neural machine translation architecture}
		\label{fig:fairseq}
\end{figure}

We use the Fairseq \cite{DBLP:journals/corr/abs-1904-01038} implementation
of Transformer \cite{DBLP:journals/corr/VaswaniSPUJGKP17} sequence-to-sequence
models to build all our models.
The architecture of a Transformer model is shown in
Fig. \ref{fig:fairseq}. The model consists of two components: an encoder and
a decoder. Each contains of set of stacked layers composed primarily
of multi-head attention sublayers that feed into feedforward sublayers. The attention sublayers in the
encoder only attend to the input sequence (and features derived from it).
The decoder can attend to the partial sequence of generated tokens
and is masked to prevent it from attending to future tokens. The decoder also has
attention on the output of the encoder stack.
The heads in an attention sublayer are able to form independent representations that may attend to different positions.

The architecture we used had 6-layer encoder and decoder stacks, 8
attention heads, embedding dimension 512, and feedforward network
embedding dimension 2048.  Training is performed on a parallel corpus
using stochastic gradient descent (SGD).

\subsection{Byte Pair Encoding}
\label{sec:bpe}

We follow the MT practice of preprocessing the training data.
Byte Pair Encoding (BPE) in the context of dealing with rare words is
described by Sennrich et al. \cite{DBLP:journals/corr/SennrichHB15}.
BPE encodes the most frequent character bigrams as unseen
unigrams and the process is repeated until a stopping point is
reached.
BPE is a form of data compression. It is also a way to deal with
out-of-vocabulary words by attempting to break them down into
component parts. Table \ref{tab:ex1} shows an example of the
BPE-processed data. The @@ symbol is used to indicate where BPE has
divided a word into subwords.
In \cite{amazon}
there is a detailed analysis of using BPE and different data sizes for a
normalization task.
BPE encoding is used for all the models we build.

	\begin{table}[bt]
		\centering
		\footnotesize

		\begin{tabular}{| l | l |}
			\hline
			before  & Most Watched TV Show (Scripted): “Bonanza”  \\
                        \hline
			after  & Most Wat@@ ched TV Show (S@@ crip@@ ted@@ ): \\
                        & “@@ Bon@@ an@@ za@@ ” \\
			\hline
		\end{tabular}
		\caption{Effect of BPE on data}
		\label{tab:ex1}
	\end{table}

\subsection{Dual model}

For the dual model case we divide the problem of translating text into two parts
(1) normalization (with or without punctuation)
and, (2) pronunciation

The system is trained on several million sentences in
parallel (details below).
Training
is carried out using standard Fairseq recipes, with a training set,
validation set and test set.

\subsection{Single model}

For the single model case we combine normalization and pronunciation
into one model. Training
is carried out in a similar fashion to the dual models.

\subsection{Splicing}

For either the single or dual model case it is important to have a
strategy for dealing with arbitrarily long sentences. We take a
straightforward approach of dividing longer sentences into multiple
overlapping parts of length 25 words, with an overlap of 10 words,
without regard to sentence boundaries or syntax.  To produce the final
output we align overlapping output sequences by maximizing the word
level agreement in the overlap.

\section{EXPERIMENTS}
\label{sec:experiments}

To test how easy it was to bring up a FE for a new locale, we built
models for 18 different locales and compared model
performance across locales.

For each locale our training data consists of roughly 5 million
sentences. The data was collected by web crawling and processed to
extract sentences. No specific limitations were put on the form of
these data.

Next, these sentences were input to a working production synthesizer
and intermediate and final processed forms of the data were extracted
to give a database of parallel sentences in ``unnormalized'',
``normalized'' and ``phone sequence'' forms.

\begin{table}[ht]
\centering
\begin{tabular}{|l|l|l|}
\hline
Model & Input & Output \\ [0.5ex]
\hline
Normalization & unnormalized & normalized \\
Pronunciation & normalized & phone sequence \\
Combined & unnormalized & phone sequence \\

\hline
\end{tabular}
\caption{Source and Target Data for each S2S Model}
\label{tab:src-tgt}
\end{table}

The source and target data for each type of model is shown in Table \ref{tab:src-tgt}.
Each of these sets formed the initial basis for training a model.
From the data sets described, and for each locale of interest we then
trained a model using the method outlined in Section \ref{subsec:fairseq} above.
For the specific models described here, no tokenization was carried
out on the data prior to the BPE step. We used a joint BPE, with a
codebook of 32k pairs, and 16-GPU Fairseq configuration in our
experiments. We held out 10,000 sentences for validation, and 10,000
sentences for testing. None of the validation or test sentences were
contained in the training data.

Speed was not considered here. In general generating an encoding can
be somewhat slow, but encoding and decoding thereafter is not
expensive.

\subsection{Listening tests}

Testing the output quality of the FE models in a complete synthesis
system presents certain complexities. First, text differences compared
with the teacher system are infrequent and usually minor.  Second, the
models form part of a Neural E2E system, with a Neural Backend
(BE). When comparing with a Unit Selection production system any
listening tests will inevitably reflect the influence of the BE.
Nevertheless, to find out whether the FE models are able to provide
all the information necessary for a production scenario we ran
listening tests comparing the completely E2E system with the production system for
several locales.

\section{RESULTS}
\label{sec:results}

\subsection{Accuracy results}
\label{sec:accuracy}

Two measures were used to assess accuracy on the held out test set:
BLEU \cite{papineni} and chrF3 \cite{popovic}.

We observed that given a high-quality database the
training will reproduce the patterns in the data to a high degree of
accuracy.

Results for the dual model case are given in Table
\ref{tab:twomodelsspliced}. Generally the accuracy was lower, but
still reasonable for most synthesis cases. For longer sentences the
test outputs are created by splicing multiple shorter outputs.

Accuracy variation among locales may be reflective of the structure of the language
or the extent to which the process (BPE and Fairseq) aligns with
language. We did observe some correlation with the quality of
the training data.

\begin{table}[ht]
\centering
\begin{tabular}{|l|r|r|r|r|r|r}
\hline
Locale & \multicolumn{2}{c|}{Normalization} & \multicolumn{2}{c|}{Pronunciation} \\ [0.5ex]
 & BLEU & chrF3 & BLEU & chrF3 \\ [0.5ex]
\hline
en-US & 99.69 & 0.9991 & 97.09 & 0.9926 \\
es-ES & 99.79 & 0.9990 & 99.88 & 0.9996 \\
it-IT & 99.80 & 0.9994 & 99.71 & 0.9991 \\
pt-PT & 99.85 & 0.9993 & 99.68 & 0.9992 \\
fr-FR & 99.70 & 0.9991 & 99.52 & 0.9985 \\
sv-SE & 99.10 & 0.9934 & 99.34 & 0.9970 \\
nl-NL & 98.13 & 0.9855 & 98.62 & 0.9925 \\
en-AU & 99.60 & 0.9870 & 98.91 & 0.9882 \\
de-DE & 99.80 & 0.9877 & 95.87 & 0.9895 \\
ru-RU & 99.00 & 0.9942 & 99.10 & 0.9964 \\
da-DK & 97.07 & 0.9915 & 97.94 & 0.9894 \\
en-IN & 99.15 & 0.9969 & 99.51 & 0.9974 \\
nb-NO & 93.93 & 0.9808 & 96.22 & 0.9853 \\
en-ZA & 98.20 & 0.9855 & 98.02 & 0.9865 \\
en-IE & 97.72 & 0.9810 & 97.65 & 0.9833 \\
tr-TR & 94.20 & 0.9763 & 98.14 & 0.9853 \\
en-GB & 83.66 & 0.9005 & 99.56 & 0.9975 \\
pt-BR & 79.10 & 0.6585 & 95.86 & 0.9673 \\

\hline
\end{tabular}
\caption{Testing Accuracy -- dual model}
\label{tab:twomodelsspliced}
\end{table}

For the single model case refer to Table
\ref{tab:onemodelspliced}. Generally the accuracy was lower, but still
very reasonable for most synthesis cases.  This table also illustrates
the significant performance boost achieved by using the splicing
technique.

\begin{table}[ht]
\centering
\begin{tabular}{|l|r|r|r|r|}
\hline
Locale & \multicolumn{2}{c|}{Combined} & \multicolumn{2}{c|}{Combined, Spliced} \\ [0.5ex] %
 & BLEU & chrF3 & BLEU & chrF3 \\ [0.5ex] %
\hline
de-DE & 92.01 & 0.9484 & 94.82 & 0.9782 \\
en-US & 92.94 & 0.9428 & 96.84 & 0.9822 \\
es-ES & 91.51 & 0.9246 & 99.54 & 0.9969 \\
nl-NL & 94.42 & 0.9509 & 97.41 & 0.9826 \\
ru-RU & 94.46 & 0.9558 & 98.48 & 0.9919 \\
sv-SE & 97.39 & 0.9789 & 98.41 & 0.9891 \\

\hline
\end{tabular}
\caption{Testing Accuracy -- single model, unspliced and spliced}
\label{tab:onemodelspliced}
\end{table}

We analyzed modeling errors, particularly for the text normalization
component. For en-US, of our 10,000 test sentences, 127 differed from the
baseline reference. We divided those 127 into the categories shown in
Table \ref{tab:errors}. None of the errors appeared completely random, most of
the differences were minor. Cases labeled ``punctuation'', for example
often involved hyphens being absent or present. Cases labeled ``2nd
lang'' contained substantial amounts of other language text.

\begin{table}[ht]
\centering
\begin{tabular}{|r|l|r|}
\hline
Instances & Type & Percentage  \\ [0.5ex]
\hline
 21 & better &  16\% \\
 11 & equal  &   9\% \\
 39 & punctuation  &  31\% \\
  4 & 2nd lang &   3\% \\
 52 & worse  &  41\% \\
\hline
\end{tabular}
\caption{Breakdown by type of differences in en-US text normalization}
\label{tab:errors}
\end{table}

Some example differences are shown in Table \ref{tab:errorexamples}.
Some of the differences appear to be because the MT generalizes better than the hand-crafted baseline.
The baseline system is the one used to generate the training data.

\begin{table}[ht]
\centering
\begin{tabular}{|r|l|}
\hline

\hline
original &   11.40AM IST \\
baseline &   eleven forty A M ist \\
MT       &   eleven forty A M irish summer time  \\
\hline
\hline
original &    (A Yuuuuge amount of articles?),\\
baseline &    A yuuuuge amount of articles.\\
MT       &    A yuuuge amount of articles.\\
\hline
\hline
original &    a wind that stiffened to 70kmh by lunch.\\
baseline &    a wind that stiffened to seventy K M H \\
         &    by lunch\\
MT       &    a wind that stiffened to seventy kilometers\\
         &    per hour by lunch\\

\hline
\end{tabular}
\caption{Examples of differences in en-US text normalizations}
\label{tab:errorexamples}
\end{table}

BPE failed very rarely. For example, failure happened for only 2
tokens in 5 million sentences (80 million+ tokens) for en-US
normalization. These tokens were replaced by $\langle$unk$\rangle$.

When run on a test set of examples of interest in a voice assistant
context we found some cases where our original data was too sparse to
model accurately. The most frequent such case was dealing with extremely
large numbers but there were also cases, for example, where
mathematical symbols were not rendered correctly because they were
underrepresented in the training data.

It is worth noting that the models for fr-* locales were able to model
liaisons without the need for adapting the framework, evidence in
support of the generality of the approach.

\subsection{Listening test}

We also incorporated the models as part of a full TTS
pipeline, ran them end to end and performed listening tests. The S2S FE
components gave essentially identical results for these real world
testing cases. (The cases are generally less challenging than our held
out test set).

Table \ref{tab:listening} shows partial results from various
subjective listening tests measuring naturalness for an E2E system
compared to a production system. Any quality improvements are
reflective of the BE.

\begin{table}[ht]
\centering
\begin{tabular}{|l|l|l|}
\hline
Locale & E2E vs. Prod. & Mean Opinion Scores \\
       & & (if significantly different) \\
\hline
 es-MX & equivalent & \\
 it-IT & better     & 4.29 vs 3.66 \\
 sv-SE & better     & \\
 ru-RU & better     & 4.10 vs 3.66 \\
 de-DE & equivalent & \\
 en-AU & better     & 4.30 vs 4.17 \\
 en-US & equivalent & \\
 en-GB & better     & 4.38 vs 4.20 \\
 fr-FR & better     & 4.48 vs 4.41 \\
\hline
\end{tabular}
\caption{Partial summary of listening experiments, focusing on Neural E2E system}
\label{tab:listening}
\end{table}

\section{CONCLUSIONS}
\label{sec:concludions}

The main contributions of this paper is a general
framework where S2S models can replace the FE of an existing
rule-based TTS. The existing system is used as a teacher to help train
the models, providing normalized and pronunciation forms for a large
database of unnormalized sentences. These parallel data are then used as
input data for S2S training.

For the configurations we studied, the dual model gives the better
performance, however the single model is smaller overall and has
better processing characteristics, and in terms of quality is
comparable to the dual model.

Our approach to pronunciations does not rely explicitly on a lexicon
or isolated word pronunciation modeling, but instead provides a general
language-independent framework for dealing with pronunciations in context.

For inference we introduced the language-independent idea of cutting
input data into chunks and splicing the outputs back together. This
both improved the accuracy of our models and allows the synthesis of
arbitrary-length sentences.

To demonstrate the scalability and generality of our approach, we
presented a large-scale study where we trained models for 18 locales
and measured high accuracies. We also tested our models in a full
synthesis context against a production system. Under testing, the FE
models were found to be robust and accurate.

\vfill\pagebreak

\bibliographystyle{IEEEbib}
\bibliography{strings,refs}

\end{document}